\documentclass[sigconf]{acmart}

\usepackage{amsmath,amsfonts,bm}

\def\eqref#1{equation~\ref{#1}}

\def\1{\bm{1}}

\def\vc{{\bm{c}}}

\def\vh{{\bm{h}}}

\def\vp{{\bm{p}}}

\def\vr{{\bm{r}}}
\def\vs{{\bm{s}}}
\def\vt{{\bm{t}}}
\def\vu{{\bm{u}}}
\def\vv{{\bm{v}}}

\def\vx{{\bm{x}}}

\def\mC{{\bm{C}}}

\def\mM{{\bm{M}}}

\def\mV{{\bm{V}}}

\def\mX{{\bm{X}}}

\DeclareMathAlphabet{\mathsfit}{\encodingdefault}{\sfdefault}{m}{sl}
\SetMathAlphabet{\mathsfit}{bold}{\encodingdefault}{\sfdefault}{bx}{n}

\def\gD{{\mathcal{D}}}
\def\gE{{\mathcal{E}}}

\def\gG{{\mathcal{G}}}

\def\gL{{\mathcal{L}}}

\def\gN{{\mathcal{N}}}

\def\gR{{\mathcal{R}}}

\newcommand{\R}{\mathbb{R}}

\newcommand{\softmax}{\mathrm{softmax}}
\newcommand{\sigmoid}{\sigma}

\usepackage{url}
\usepackage{graphicx}
\usepackage{color}
\usepackage{multirow}
\usepackage{subfigure}

\DeclareMathOperator*{\mlp}{MLP}
\DeclareMathOperator*{\transformerenc}{Transformer-Enc}
\DeclareMathOperator*{\pool}{Pool}
\DeclareMathOperator*{\distance}{Distance}

\DeclareMathOperator*{\rescale}{Rescale}
\DeclareMathOperator*{\std}{Std}
\DeclareMathOperator*{\mean}{Mean}

\AtBeginDocument{%
  \providecommand\BibTeX{{%
    \normalfont B\kern-0.5em{\scshape i\kern-0.25em b}\kern-0.8em\TeX}}}
    
\copyrightyear{2021}
\acmYear{2021}
\setcopyright{iw3c2w3} 

\acmConference[WWW '21]{Proceedings of the Web Conference 2021}{April 19--23, 2021}{Ljubljana, Slovenia}
\acmBooktitle{Proceedings of the Web Conference 2021 (WWW '21), April 19--23, 2021, Ljubljana, Slovenia}
\acmPrice{}
\acmDOI{10.1145/3442381.3450043}
\acmISBN{978-1-4503-8312-7/21/04}

\begin{document}
\title{
Structure-Augmented Text Representation Learning for Efficient\\ Knowledge Graph Completion}

\author{Bo Wang$^{1}$, Tao Shen$^{2}$, Guodong Long$^{2}$, Tianyi Zhou$^{3}$, Ying Wang$^{4*}$, Yi Chang$^{1,5*}$}
\thanks{~~*Joint Corresponding Author}

\affiliation{
\institution{$^1$School of Artificial Intelligence, Jilin University\\ 
$^2$Australian AI Institute, School of CS, FEIT, University of Technology Sydney \\
$^3$Paul G. Allen School of Computer Science \& Engineering, University of Washington, Seattle \\
$^4$College of Computer Science and Technology, Jilin University \\
$^5$International Center of Future Science, Jilin University
}
}
\email{bowang19@mails.jlu.edu.cn, tao.shen@student.uts.edu.au, guodong.long@uts.edu.au,} 
\email{tianyizh@uw.edu, wangying2010@jlu.edu.cn, yichang@jlu.edu.cn}

\renewcommand{\authors}{Bo Wang, Tao Shen, Guodong Long, Tianyi Zhou, Ying Wang, Yi Chang}
\renewcommand{\shortauthors}{B. Wang et al.}

\begin{abstract}
Human-curated knowledge graphs provide critical supportive information to various natural language processing tasks, but these graphs are usually incomplete, urging auto-completion of them (a.k.a. knowledge graph completion). 
Prevalent graph embedding approaches, e.g., TransE, learn structured knowledge via representing graph elements (i.e., entities/relations) into dense embeddings and capturing their triple-level relationship with spatial distance. 
However, they are hardly generalizable to the elements never visited in training and are intrinsically vulnerable to graph incompleteness. 
In contrast, textual encoding approaches, e.g., KG-BERT, resort to graph triple's text and triple-level contextualized representations. They are generalizable enough and robust to the incompleteness, especially when coupled with pre-trained encoders. But two major drawbacks limit the performance: (1) high overheads due to the costly scoring of all possible triples in inference, and (2) a lack of structured knowledge in the textual encoder. 
In this paper, we follow the textual encoding paradigm and aim to alleviate its drawbacks by augmenting it with graph embedding techniques -- a complementary hybrid of both paradigms. 
Specifically, we partition each triple into two asymmetric parts as in translation-based graph embedding approach, and encode both parts into contextualized representations by a Siamese-style textual encoder. 
Built upon the representations, our model employs both deterministic classifier and spatial measurement for representation and structure learning respectively. 
It thus reduces the overheads by reusing graph elements' embeddings to avoid combinatorial explosion, and enhances structured knowledge by exploring the spatial characteristics. 
Moreover, we develop a self-adaptive ensemble scheme to further improve the performance by incorporating triple scores from an existing graph embedding model. In experiments, we achieve state-of-the-art performance on three benchmarks and a zero-shot dataset for link prediction, with highlights of inference costs reduced by 1-2 orders of magnitude compared to a sophisticated textual encoding method. 
\end{abstract}


\maketitle

\section{Introduction} \label{sec:intro}

Knowledge graph (KG) is a ubiquitous format of knowledge base (KB). It is structured as a directed graph whose vertices and edges respectively stand for entities and their relations. It is usually represented as a set of triples in the form of (\textit{head entity, relation, tail entity}), or (\textit{h, r, t}) for short. 
KGs as supporting knowledge play significant roles across a wide range of natural language processing (NLP) tasks, such as dialogue system \cite{he2017dialogue}, information retrieval \cite{xiong2017IR}, recommendation system \cite{zhang2016RecSys}, etc. 
However, human-curated knowledge graphs usually suffer from incompleteness \cite{socher2013kgc}, inevitably limiting their practical applications. 
To mitigate this issue, knowledge graph completion (KGC) aims to predict the missing triples in a knowledge graph. In this paper, we particularly target \textit{link prediction} task for KGC, whose goal is to predict the missing \textit{head} (\textit{tail}) given the \textit{relation} and \textit{tail} (\textit{head}) in a triple.

It is noteworthy that KG \cite{db/freebase, db/YAGO, db/Wikidata} is usually at a scale of billions and the number of involved entities is up to millions, so most graph neural networks (e.g., GCN \cite{GCN}) operating on the whole graph are not scalable in computation.
Thus, approaches for KGC often operate at triple level, which can be grouped into two paradigms, i.e., \textit{graph embedding} and \textit{textual encoding} approaches. 

\textit{Graph embedding} approaches attempt to learn the representations for graph elements (i.e., entities/relations) as low-dimension vectors by exploring their structured knowledge in a KG. 
Typically, they directly exploit the spatial relationship of the three embeddings in a triple to learn structured knowledge, which can be classified into two sub-categories. 
(1) \emph{Translation-based} approaches, e.g., TransE \cite{TransE} and RotatE \cite{sun2019rotate}, score the plausibility of a triple by applying a translation function to the embeddings of head and relation, and then measuring how close the resulting embedding to the tail embedding, i.e., $-||g(\vh, \vr) - \vt||_p$; 
And (2) \emph{semantic matching} approaches, e.g., DistMult \cite{yang2014DistMult} and QuatE \cite{QuatE}, derive the plausibility of a graph triple via a matching function that directly operates on the triple, i.e., $f(\vh, \vr, \vt)$. 
Despite their success in structure learning, they completely ignore contextualized information and thus have several drawbacks: 
(1) The trained models are not applicable to entities/relations unseen in training; 
And (2) they are intrinsically vulnerable to the graph incompleteness.
These drawbacks severely weaken their generalization capability and prediction quality.

\textit{Textual encoding}\footnote{``\textit{Textual encoding}'' in this paper refers to capturing contextualized information across entities and relations in a triple \cite{yao2019kgbert},
despite previous works (as detailed in \S \ref{subsec:compared_text-based}) using the text of a stand-alone entity/relation to enhance corresponding graph embedding.} approaches, e.g., KG-BERT \cite{yao2019kgbert}, predict the missing parts for KGC using the contextualized representation of triples' natural language text.  
The text can refer to textual contents of entities and relations (e.g., their names or descriptions). 
Coupled with pre-trained word embedding \cite{mikolov2013distributed,pennington2014glove} or language model \cite{BERT,RoBERTa}, the textual encoder can easily generalize to unseen graph elements and is invulnerable to graph incompleteness issue. 
However, they are limited by two inherent constraints: 
(1) Applying textual encoder to link prediction requires costly inference on all possible triples, causing a combinatorial explosion; 
(2) The textual encoder is incompetent in structure learning, leading to a lack of structured knowledge and the entity ambiguity problem \cite{Cucerzan2007disambiguation}.

\begin{figure}
    \subfigure{\includegraphics[width=0.235\textwidth]{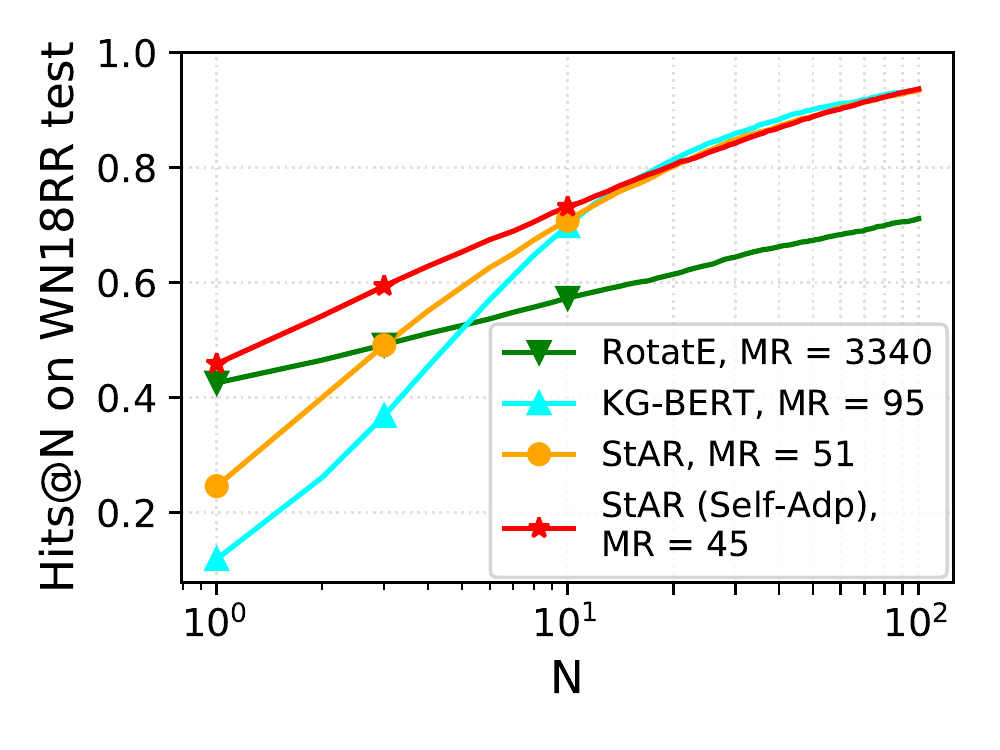}}
    \subfigure{\includegraphics[width=0.235\textwidth]{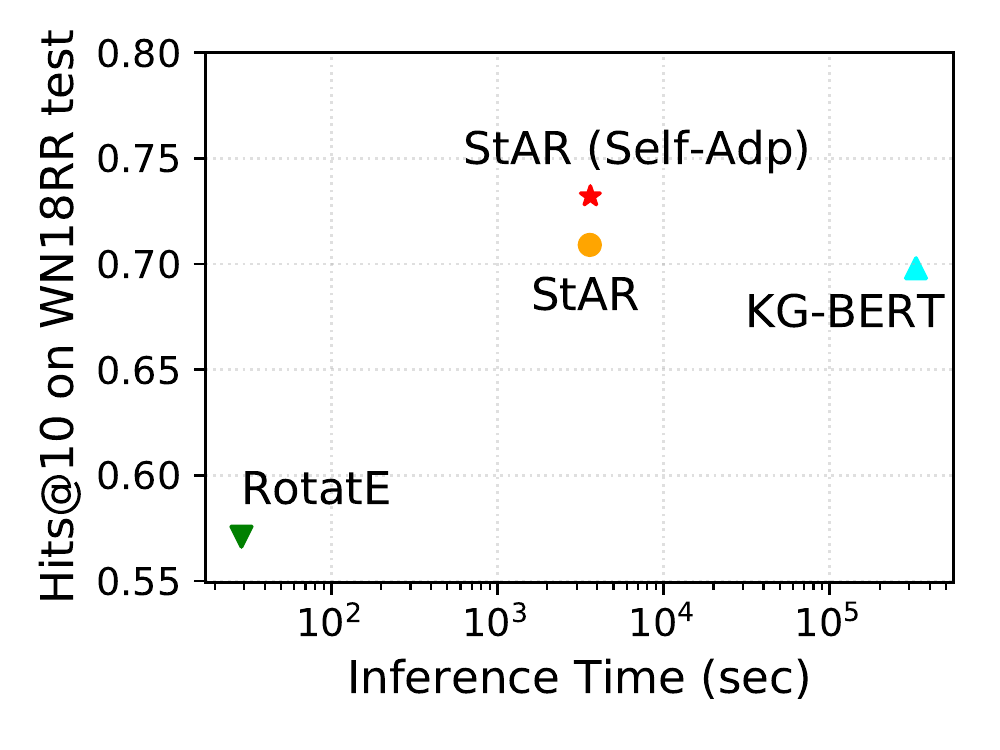}}
    \vspace{-9mm}
	\caption{\small Summary comparisons on WN18RR test. RotatE and KG-BERT are state-of-the-art approaches of \textit{graph embedding} and \textit{textual encoding} paradigms respectively. 
	StAR is our model, and ``StAR (Self-Adp)'' is our model plus our proposed self-adaptive ensemble.
	}
	\label{fig:intro_illustration} 
\end{figure}

The experiments of these two paradigms also reflect their individual Pros and Cons. 
As shown in Figure \ref{fig:intro_illustration} (left), KG-BERT achieves a high Hits@$N$ (i.e., Top-$N$ recall) when $N$ is slightly large, but fails for small $N$ due to entity ambiguity problem. In contrast, RotatE achieves a high Hits@$1$/$@3$ since it purely learns from structured knowledge without exposure to the ambiguity problem. But it still underperforms due to a lack of text contextualized information. 
And in Figure \ref{fig:intro_illustration} (right), although KG-BERT outperforms RotatE, it requires much higher overheads due to combinatorial explosion.

Therefore, it is natural to integrate both the contextualized and structured knowledge into one model, while in previous works they are respectively achieved by textual encoding and graph embedding paradigms. 
To this end, we start with a textual encoding paradigm with better generalization and then aim at alleviating its intrinsic drawbacks, i.e., overwhelming overheads and insufficient structured knowledge. 
Specifically, taking the inspiration from translation-based graph embedding approach (e.g., TransE), we first partition each triple into two parts: one with a combination of \textit{head} and \textit{relation}, while the other with \textit{tail}. Then, by applying a Siamese-style textual encoder to their text, we encode each part into separate contextualized representation. 
Lastly, we concatenate the two representations in an interactive manner \cite{reimers2019sentbert} to form the final representation of the triple and train a binary neural classifier upon it. 
In the meantime, as we encode the triple by separated parts, we can measure their spatial relations like translation function \cite{TransE,sun2019rotate} and then conduct a structure learning using contrastive objective. 

Consequently, on the one hand, our model can re-use the same graph elements' embeddings for different triples to avoid evaluating the combinatorial number of triples required in link prediction. On the other hand, it also augments the textual encoding paradigm by modeling structured knowledge, which is essential to graph-related tasks. 
In addition, our empirical studies on link prediction show that introducing such structured knowledge can effectively reduce false positive predictions and help entity disambiguation. 
As shown in Figure~\ref{fig:intro_illustration}, our model improves the KG-BERT baseline in both performance and efficiency, but given a small $N$ (e.g., $\leq 2$), the performance is not that satisfactory. 
Motivated by this, we propose a self-adaptive ensemble scheme that incorporates our model's outputs with the triple scores produced by an existing graph embedding model (e.g., RotatE). Thereby, we can benefit from the advantages of both the graph embedding and textual encoding. Hence, as shown in Figure~\ref{fig:intro_illustration}, our model plus the proposed self-adaptive ensemble with RotatE achieves more.
Our main contributions are:
\begin{itemize}
    \item We propose a hybrid model of \textit{textual encoding} and \textit{graph embedding} paradigms to learn both contextualized and structured knowledge for their mutual benefits: A Siamese-style textual encoder generalizes graph embeddings to unseen entities/relations, while augmenting it with structure learning contributes to entity disambiguation and high efficiency. 
    \item We develop a self-adaptive ensemble scheme to merge scores from graph embedding approach and boost the performance. 
    \item We achieve state-of-the-art results on three benchmarks and a zero-shot dataset; We show a remarkable speedup (6.5h vs. 30d on FB15k-237 \cite{FB15k-237}) over recent KG-BERT \cite{yao2019kgbert}; We provide a comparative analysis of the two paradigms. 
\end{itemize}

\section{Background}  \label{sec:background}
We start this section with a formal definition of the link prediction task for KGC. 
Then, we summarize the pre-trained masked language model and its fine-tuning.
And lastly we give a brief introduction to a state-of-the-art textual encoding approach, KG-BERT \cite{yao2019kgbert}. 

\paragraph{Link Prediction.}
Formally, a KG $\gG = \{\gE, \gR\}$ consists of a set of triples (\textit{h, r, t}), where $h,t\in\gE$ are head and tail entity respectively while $r\in\gR$ is the relation between them. 
Given a head $h$ (tail $t$) and a relation $r$, the goal of link prediction is to find the most accurate tail $t$ (head $h$) from $\gE$ to make a new triple (\textit{h, r, t}) plausible in $\gG$. 
And during inference, given an incomplete triple (\textit{h, r, ?}) for example, a trained model is asked to score all candidature triples $\{(h,r,t')| t'\in\gE\}$ and required to rank the only oracle triple $(h,r,t^*)$ as high as possible. This is why the combinatorial explosion appears in a computation-intensive model defined at triple level.

\paragraph{Pre-Trained Masked Language Model.}  
To obtain powerful textual encoders, masked language models (MLMs) pre-trained on large-scale raw corpora learn generic contextualized representations in a self-supervised fashion (e.g., BERT \citep{BERT} and RoBERTa \cite{RoBERTa}). 
MLMs randomly mask some tokens and predict the masked tokens by considering their contexts on both sides. Specifically, given tokenized text $[w_1, \dots, w_n]$, a certain percentage (e.g., 15\% in \cite{BERT}) of the original tokens are then masked and replaced: of those, 80\% with special token \texttt{[MASK]}, 10\% with a token sampled from the vocabulary $\mathbb{V}$, and the remaining kept unchanged. The masked sequence of embedded tokens $[\vx_1^{(m)}, \dots, \vx_n^{(m)}]$ is passed into a Transformer encoder \cite{transformer} to produce contextualized representations for the sequence:
\begin{align}
	\mC  =  \transformerenc([\vx_1^{(m)}, \dots, \vx_n^{(m)}]) \in \mathbb{R}^{d_h\times n}. \label{eq:trans_enc}
\end{align}
The pre-training loss is defined as 
\begin{align}
	\mathcal{L}^{m} = - \dfrac{1}{|\mathcal{M}|} \sum\nolimits_{i \in \mathcal{M}} \log P(w_i|\mC_{:,i}), \label{eq:loss_mlm}
\end{align}
where $\mathcal{M}$ is the set of masked token indices, and $P(w_i|\mC_{:,i})$ is the probability of predicting the masked $w_i$.
After pre-trained, they act as initializations of textual encoders, performing very well on various NLP tasks with task-specific modules and fine-tuning \cite{BERT}.

\paragraph{KG-BERT} 
As a recent textual encoding approach \cite{yao2019kgbert} for KGC, 
instead of using embeddings of entities/relations, it scores a triple upon triple-level contextualized representation. 
Specifically, a tokenizer with a word2vec \cite{mikolov2013distributed,pennington2014glove} first transforms the text $x$ of each entity/relation to a sequence of word embeddings $\mX = [\vx_1, \dots, \vx_{n}] \in \mathbb{R}^{d\times n}$. 
So, the text of a triple ($x^{(h)}$, $x^{(r)}$, $x^{(t)}$) can be denoted as ($\mX^{(h)}$, $\mX^{(r)}$, $\mX^{(t)}$). 
Then, KG-BERT applies the Transformer encoder \cite{transformer} to a concatenation of the \textit{head}, \textit{relation} and \textit{tail}. The encoder is initialized by a pre-trained MLM, BERT, and the concatenation is 
$\tilde \mX = [\vx^{\texttt{[CLS]}}, \mX^{(h)}, \vx^{\texttt{[SEP]}}, \mX^{(r)}, \vx^{\texttt{[SEP]}}, \mX^{(t)},  \vx^{\texttt{[SEP]}}]$, where \texttt{[CLS]} and \texttt{[SEP]} are special tokens defined by \citet{BERT}.
Based on this, KG-BERT produces a contextualized representation $\vc$ for the entire triple, i.e.,
\begin{equation}
\vc = \pool(\transformerenc(\tilde \mX)),
\end{equation}
where $\pool(\cdot)$, defined in \cite{BERT}, collects the resulting of \texttt{[CLS]} to denote a contextualized representation for the sequence. 
Next, $\vc$ is passed into a two-way classifier to determine if the triple is plausible or not. Lastly, the model is fine-tuned by minimizing a cross entropy loss. 
In inference, positive probability (i.e., confidence) of a triple is used as a ranking score. 
Such a simple approach shows its effectiveness for KGC, highlighting the significance of text representation learning. 
We thus follow this line and propose our model by avoiding combinatorial explosion and enhancing structure learning.

\section{Proposed Approach} \label{sec:approach}

\begin{figure*}[t] 
	\centering
	\includegraphics[width=0.7\textwidth]{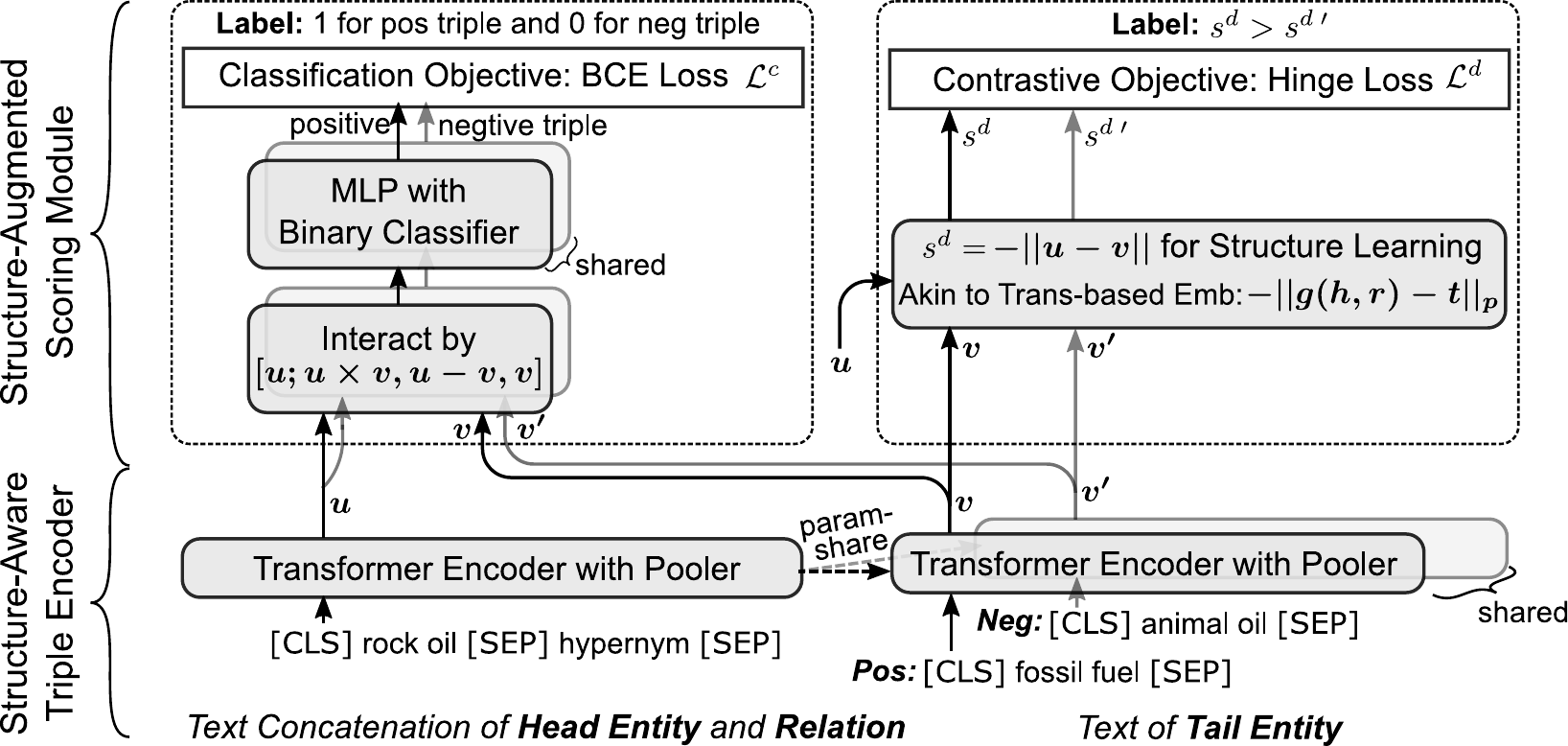}
	\caption{\small
	An overview of the proposed \textbf{St}ructure-\textbf{A}ugmented Text \textbf{R}epresentation (StAR) model for link prediction. 
	This illustration is based on a corruption of tail entity, and in the same way for the corruption of a head entity or even relation. Note that a notation whose superscript includes ``$~'~$'' denotes it is derived from a negative example, otherwise from a positive one.}
	\label{fig:model_arc} 
	\centering
\end{figure*}

In this section, we first elaborate on a structure-aware triple encoder (\S \ref{subsec:triple_enc}) and a structure-augmented triple scoring module (\S \ref{subsec:two_parallel_learn}), which compose our \textbf{St}ructure-\textbf{A}ugmented Text \textbf{R}epresentation (StAR) model to tackle link prediction for KBC (as illustrated in Figure~\ref{fig:model_arc}). 
And we provide the details about training and inference, e.g., training objectives and efficiency, in \S \ref{subsec:objectives}.
Then, we develop a self-adaptive ensemble scheme in \S \ref{subsec:adaptive model} to make the best of an existing graph embedding approach and boost the performance. 
Lastly, in \S \ref{subsec:compared_text-based}, we provide comparative analyses between our model and previous text-based approaches for graph-related tasks. 

\subsection{Structure-Aware Triple Encoding}  \label{subsec:triple_enc} 

In this subsection, we aim at encoding a graph triple into vector representation(s) in latent semantic space, with consideration of subsequent structure learning and inference speedup. 
The representation(s), similar to graph embeddings, can be fed into any downstream objective-specific module to fulfil triple scoring. 

Recently, to accelerate the inference of a deep Transformer-based model \cite{transformer,BERT} in an information retrieval (IR) task, \citet{reimers2019sentbert} adopted a two-branch Siamese architecture \cite{Chopra2015siamese} to bypass pairwise input via encoding the query and candidate separately. 
This enables pre-computing the representations for all candidates and uses a light-weight matching net \cite{reimers2019sentbert} to calculate relatedness. 
We take this inspiration to link prediction to avoid combinatorial explosion, but several open questions arise: 
(1) How to preserve contextualized knowledge across the entities and relation in a triple; 
(2) How to apply Siamese architecture to a triple with three components; 
and (3) How to facilitate structure learning in downstream modules.

These questions can be dispelled by digesting several techniques from translation-based graph embedding approaches, e.g., TransE \cite{TransE} and RotatE \cite{sun2019rotate}. 
The techniques include applying a \textit{translation function} to the embeddings of \textit{head} and \textit{relation}, 
and structure learning via exploring spatial relationship (e.g., distance) between the function's output and \textit{tail} embedding. 

Specifically, TransE and RotatE explicitly define the translation function as real vector addition and complex vector product, respectively. 
In contrast, as a textual encoding approach, we implicitly formulate the function as applying a Transformer-based encoder to a text concatenation of \textit{head} and \textit{relation}. The concatenation is:
\begin{align}
    \tilde \mX^{(h)} = [\vx^{\texttt{[CLS]}}, \mX^{(h)}, \vx^{\texttt{[SEP]}}, \mX^{(r)}, \vx^{\texttt{[SEP]}}], \label{eq:cat_left}
\end{align}
where $\vx^{\texttt{[CLS]}}$ and $\vx^{\texttt{[SEP]}}$ are embedded special tokens defined in \cite{BERT}. 
Refer to KG-BERT in \S\ref{sec:background} for the details about pre-processing. 
Then, such ``contextualizing'' translation function is defined as 
\begin{align}
\vu = \pool(\transformerenc(\tilde \mX^{(h)})), \label{eq:t-enc_left}
\end{align}
where $\transformerenc(\cdot)$ denotes the Transformer encoder consisting of multi-head self-attention layers\cite{transformer}.
We keep using the segment identifier given by \citet{BERT} to mark if a token is from an entity (i.e., $0$) or a relation (i.e., $1$). And again, $\pool(\cdot)$ collects the resulting of \texttt{[CLS]} to denote sequence-level contextualized representation. 
So, $\vu$, a contextualized representation across \textit{head} and \textit{relation}, can be viewed as the translation function's output.

On the other side, we also encode \textit{tail} by applying the Transformer encoder to its text, which is written as
\begin{align}
\vv &= \pool(\transformerenc(\tilde \mX^{(t)})), \label{eq:t-enc_right}\\
&\textit{where,}~\tilde \mX^{(t)} = [\vx^{\texttt{[CLS]}}, \mX^{(t)}, \vx^{\texttt{[SEP]}}]. \label{eq:cat_right}
\end{align}
Consequently, $\vv$, a contextualized representation of \textit{tail}, is viewed as \textit{tail} embedding. 
In our experiment, we keep the two Transformer encoders (i.e., in Eq.(\ref{eq:t-enc_left}) and (\ref{eq:t-enc_right})) parameter-tied for parameter efficiency \cite{reimers2019sentbert}. 
And it is noteworthy that the Transformer encoder can be initialized by a pre-trained language model to further boost its capacity for representation learning, which alternates between BERT \cite{BERT} and RoBERTa \cite{RoBERTa} in our experiments.

To sum up, also as answers to the questions above, 
(1) we trade off the contextualized knowledge with efficiency: keeping context across \textit{head} and \textit{relation}, while separating \textit{tail} for reusable embeddings to avoid combinatorial explosion; 
(2) we partition each triple into two asymmetric parts as in TransE: a concatenation of \textit{head} and \textit{relation}, versus \textit{tail}; 
and (3) we derive two contextualized embeddings for the two parts respectively, and aim to learn structured knowledge by exploring spatial characteristics between them.

\subsection{Structure-Augmented Scoring Module} \label{subsec:two_parallel_learn}

Given $\vu$ and $\vv$, we present two parallel scoring strategies as at the top of Figure \ref{fig:model_arc} for both contextualized and structured knowledge.

\subsubsection{Deterministic Representation Learning} \label{sec:repre_learn}

Recently, some semantic matching graph embedding approaches for KGC use deterministic strategy \cite{nguyen2017ConvKB,vu2019CapsE} to learn the representation of entities and relations. 
This strategy refers to using a binary classifier that determines if a triple is plausible or not. 
Such representation learning is especially significant to a text-based model, which has been adopted in KG-BERT for KGC and proven effective. 
But, this strategy cannot be applied to the pair of contextualized representations $\vu$ and $\vv$ produced by our Siamese-style encoder.

Fortunately, a common practice in NLP literature is to apply an interactive concatenation \cite{bowman2015snlidataset,liu2016learning,reimers2019sentbert} to the pair of representations and then perform a neural binary classifier. 
Formally, we adopt the interactive concatenation written as
\begin{align}
\vc = [\vu; \vu \times \vv; \vu - \vv; \vv],
\end{align}
where $\vc$ is used to represent the semantic relationship between the two parts of a triple. 
Then, similar to the top layer in KG-BERT, a two-way classifier is then applied to $\vc$ and produces a two-dimensional categorical distribution corresponding to the negative and positive probabilities respectively, i.e.,
\begin{align}
\vp &= P(z|\vc; \theta) \triangleq \softmax(\mlp(\vc; \theta)) \in \mathbb{R}^2, 
\end{align}
 
where $\mlp(\cdot)$ stands for a multi-layer perceptron, and $\theta$ is its learnable parameters. 
During the inference of link prediction, the $2nd$ dimension of $\vp$, i.e., the positive probability, 
\begin{align}
 s^c &= p_2 \label{eq:cls_score}
\end{align}
can serve as a score of the input triple to perform candidate ranking.

\subsubsection{Spatial Structure Learning} \label{sec:struc_learn}

In the meantime, it is possible to augment structured knowledge in the encoder by exploring the spatial characteristics between the two contextualized representations. 
Typically, translation-based graph embedding approaches conduct structure learning by measuring spatial distance. 
In particular, TransE \cite{TransE} and RotatE \cite{sun2019rotate} score a triple inversely proportional to the spatial distance between $g(\vh, \vr)$ and $\vt$, i.e., $-||g(\vh, \vr) - \vt||$. 
And structured knowledge is acquired via maximizing the score margin between a positive triple and its corruptions (i.e., negative triples). 

Here, as a triple is partitioned into two asymmetric parts by imitating the translation-based approaches, we can formulate $\vu \leftarrow f(h, r)$ and $\vv \leftarrow f(t)$, where $f(\cdot)$ denotes the textual encoder in \S\ref{subsec:triple_enc}. 
So, to acquire structured knowledge, we can score a triple by 
\begin{align}
s^d &= \!-\!\distance(f(h, r), f(t)) \triangleq \!-||\vu - \vv||, \label{eq:dist_measure}
\end{align}
where $||\cdot||$ denotes $L2$-norm, and $s^d$ is the plausible score based on the two contextualized representations, $\vu$ and $\vv$ , of a triple.

\subsection{Training and Inference} \label{subsec:objectives}

\subsubsection{Training Objectives and Inference Details} \label{subsubsec:train_infer_details}

Before presenting two training objectives, it is necessary to perform negative sampling and generate wrong triples. 
In detail, given a correct triple ${tp} = (h, r, t)$, we corrupt the triple and generate its corresponding wrong triple ${tp}'$ by replacing either the head or tail entity with another entity randomly sampled from the entities $\gE$ on $\gG$ during training, which satisfies ${tp}' \in \{(h, r, t')| t' \in \gE \wedge (h, r, t') \notin \gG \}$ or ${tp}' \in \{(h', r, t)| h' \in \gE \wedge (h', r, t) \notin \gG \}$, where $\gE$ denotes the ground set of all unique entities on $\gG$. In the remainder, a variable with superscript ``$~'~$'' means that it is derived from a negative example. 

\paragraph{Triple Classification Objective.} 
Given the resulting confidence score $s^c$ from Eq.(\ref{eq:cls_score}) in deterministic representation learning (\S\ref{sec:repre_learn}), we employ the following binary cross entropy loss to train the encoder w.r.t this objective, i.e., 
\begin{align}
    \mathcal{\gL}^c = - \dfrac{1}{|\gD|} 
    \sum_{tp \in \gD} \! \dfrac{1}{1\!\!+\!\!|\gN(tp)|} \!
    \left( \log s^{c} \! + \!\!\!\! \sum_{tp' \in \gN(tp)} \log(1 \! - \! s^{c}~\!') \right),  \label{eq:cls_loss} 
\end{align}
where $\gD$ denotes the training set containing only correct triples, $\gN(tp)$ denotes a set of wrong triples generated from $tp$, 
$s^{c}$ denotes positive probability of $tp$ and $(1 - s^{c}~\!')$ denotes negative probability of the wrong triple $tp'$. 
We empirically find such representation learning using the deterministic strategy is critical to the success of textual encoding KGC, consistent with previous works \cite{yao2019kgbert,reimers2019sentbert}. 

However, $s^c$ might not contain sufficient information for ranking during inference since it is only the confidence for a single triple's correctness that does not take other triple candidates into account. This may cause inconsistency between the model's training and inference. 
To compromise, loss weights in Eq.(\ref{eq:cls_loss}) must be imbalanced between positive and negative examples, i.e., $|\gN(tp)| \gg 1$, to distinguish the only positive triple among hundreds of thousands of corrupted ones during inference. 
Nonetheless, over-confident false positive predictions for a corruption (i.e., assigning a corrupted triple with $s^c \rightarrow 1.0$) still frequently appear to hurt the performance. These thus emphasize the importance of structure learning. 

\paragraph{Triple Contrastive Objective.} 
Given the distance-based score $s^d$ from Eq.(\ref{eq:dist_measure}) in spatial structure learning (\S\ref{sec:repre_learn}), we also train the encoder by using a contrastive objective. 
The contrastive objective considers a pairwise ranking between a correct triple and a wrong triple, where the latter is corrupted from the former by negative sampling. 
Formally, let $s^d$ denote the score derived from a positive triple $tp$ and $s^{{d}~\!'}$ denote the score derived from a wrong triple $tp~\!'$, we define the loss by using a margin-based hinge loss function, i.e., 
\begin{align}
    \mathcal{\gL}^d = \dfrac{1}{|\gD|} 
    \sum_{tp \in \gD} \dfrac{1}{|\gN(tp)|} 
    \sum_{tp~\!' \in \gN(tp)} \max(0, \lambda - s^{d} + s^{{d}~\!'}). \label{equ:hingeloss}
\end{align}
In experiments, we qualitatively reveal that structure learning is significant to reducing false positive and disambiguating entities, and pushes our model to produce more reliable ranking scores.

\paragraph{Training and Inference Strategies.}
The loss $\gL$ to train the StAR is a sum of the two losses defined in Eq.(\ref{eq:cls_loss}) and Eq.(\ref{equ:hingeloss}), i.e., 
\begin{align}
\mathcal{\gL} = \mathcal{\gL}^c + \gamma \mathcal{\gL}^d, \label{eq:loss_total}
\end{align}
where $\gamma$ is the weight.
After optimizing StAR w.r.t $\gL$, $s^c$, $s^d$ or their integration can be used as ranking basis during inference. We will present a thorough empirical study of the possible options of ranking score based on $s^c$ and $s^d$ in \S\ref{subsec:ablation}.

\begin{table}[t] \small
\caption{\small Inference efficiency comparison. $L$ is the length of triple text. $|\gE|$ and $|\gR|$ are the numbers of all unique entities and relations in the graph respectively. Usually, $|\gE|$ exceeds hundreds of thousands and is much greater than $|\gR|$. }
\setlength{\tabcolsep}{1pt}
    \centering
    \begin{tabular}{l|ccc}
        \hline
        \textbf{Inference on}  & \textbf{Method}  & \textbf{Complexity}  & \textbf{Speed up} \\
        \hline
        \multirow{2}{*}{One Triple}   & KG-BERT   & $O(L^2|\gE|)$  & \multirow{2}{*}{$\sim 4\times$} \\
         ~      & StAR     & $O((L/2)^2(1+|\gE|))$  &  \\
         \hline
         \multirow{2}{*}{Entire Graph} & KG-BERT   & $O(L^2|\gE|^2|\gR|)$  & \multirow{2}{*}{$\sim 4|\gE| \times$}\\
         ~      & StAR     & $O((L/2)^2|\gE|(1+|\gR|))$   &  \\
         \hline
    \end{tabular}
    \label{tab:efficiency}
\end{table}

\subsubsection{Model Efficiency.} \label{subsubsec:efficiency}

In the following, we analyze why our proposed model is significantly faster than its baseline, KG-BERT.

\paragraph{Training Efficiency.} 
As overheads are dominated by the computations happening inside the Transformer encoder, we focus on analyzing the complexity of computing the contextualized embeddings by the encoder. 
In practice, the sequence lengths of the two asymmetric parts of a triple are similar because the length of an entity's text is usually much longer than a relation's text, especially when the entity description is included \cite{xiao2017ssp,yao2019kgbert}. 
Hence, Siamese-style StAR is $2\times$ faster than KG-BERT in training as the complexity of Transformer encoder grows quadratically with sequence length.

\paragraph{Inference Efficiency.}
Similarly, we also focus on analyzing the overheads used in the encoder during inference. 
As shown in Table \ref{tab:efficiency}, we list the complexities of both KG-BERT baseline and proposed StAR, and analyze the acceleration in two cases. 
In practice, on the test set of a benchmark, our approach, without combinatorial explosion, is faster than KG-BERT by two orders of magnitude.

\subsection{Self-Adaptive Ensemble Scheme} \label{subsec:adaptive model}
StAR improves previous textual encoding approaches by introducing structure learning. It reduces those overconfident but false positive predictions and mitigates the entity ambiguity problem. 
However, compared to graph embedding operating at entity or relation level, our StAR based on the text inherently suffers from entity ambiguity. 
Fortunately, combining textual encoding with graph embedding paradigms can provide a remedy: 
Despite entity ambiguity existing, a textual encoding approach achieves a high recall in top-$k$ with slightly large $k$ (e.g., $k>5$), whereas a graph embedding approach can then precisely allocate the correct one from the $k$ candidates due to robustness to ambiguity. Note, $k \ll |\gE|$.
To the best of our knowledge, this is the first work to ensemble the two paradigms for mutual benefits. 
Surprisingly, simple averaging of the scores from the two paradigms significantly improves the performance. 
This motivates us to take a step further and develop a self-adaptive ensemble scheme. 

Given an incomplete triple (i.e., (\textit{h, r, ?}) or (\textit{?, r, t})), we aim to learn a weight $\alpha\in[0,1]$ to generate the final triple-specific score: 
\begin{align}
s^{(sa)} &= \alpha\times s^{(tc)} + (1 - \alpha)\times s^{(ge)},
\end{align}
where $s^{(tc)}$ is derived from StAR and $s^{(ge)}$ is derived from RotatE \cite{sun2019rotate}. Since $s^{(tc)}=s^c\in[0,1]$ from Eq.(\ref{eq:cls_score}) is normalized, we rescale all candidates' scores of RotatE into $[0,1]$ to obtain $s^{(ge)}$.
Specifically, for an incomplete triple, we first take the top-$k$ candidates ranked by StAR and fetch their scores from the two models, which are denoted as $\vs^{(tc)}\in[0,1]^k$ and $\vs^{(ge)}\in[0,1]^k$ respectively. 
Then, we set an \textit{unseen indicator} to force $\alpha=1$ if an unseen entity/relation occurs in the incomplete triple. 
Next, to learn a triple-specific $\alpha$, we build an MLP based upon two kinds of features: \textit{ambiguity degree} $\vx^{(ad)}$ and \textit{score consistency} $\vx^{(sc)}$. 
Particularly, the ambiguity degree $\vx^{(ad)} \triangleq [\std(\mV); \mean(\mM)]$ where ``$\std(\mV\in\R^{d\times k})\in\R^{d}$'' is the standard deviation of the top-$k$ entities' representations, and ``$\mean(\mM\in\R^{k\times 100})\in\R^{k}$'' averages the largest 100 cosine similarities between each candidate and all entities in $\gE$. 
Note each entity is denoted by its contextualized representation from Eq.(\ref{eq:t-enc_right}). And, the score consistency $\vx^{(sc)}\triangleq[|\vs^{(tc)}-\vs^{(ge)}|,\vs^{(tc)}+\vs^{(ge)},\vs^{(tc)},\vs^{(ge)}]$. 
Lastly, we pass the features into an MLP with activation $\sigmoid$, i.e., 
\begin{align}
    \alpha = \sigmoid(\mlp([\vx^{(ad)};\vx^{(sc)}]; \theta^{(\alpha)}))\in[0,1]. \label{eq:self_adp_alpha}
\end{align}

In training, we fix the parameters of both our model and RotatE, and only optimize $\theta^{(\alpha)}$ by a margin-based hinge loss. 
In inference, we use the resulting $s^{(sa)}$ to re-rank the top-$k$ candidates while keep the remaining unchanged. 
In experiments, we evaluated two variants of StAR: 
(1) \textbf{StAR (Ensemble)}: $k \leftarrow \infty$ and $\alpha \leftarrow 0.5$, equivalent to score average, as our ensemble baseline. 
(2) \textbf{StAR (Self-Adp}): $k \leftarrow 1000$ and $\alpha$ is learnable.

\subsection{Compared to Prior Text-Based Approach } \label{subsec:compared_text-based}
Sharing a similar motivation, some previous approaches also use textual information to represent entities and relations. 
However, they are distinct from textual encoding approaches like KG-BERT or StAR and can be coarsely categorized into two groups:

\paragraph{Stand-alone Embedding.}
These approaches \cite{socher2013kgc,McIlraith2018open} directly replace an entity/relation embedding in graph with its text representation. The representation is derived from applying a shallow encoder (e.g., CBoW and CNN) to text, regardless of contextual information across entities and relations. 
But, deep contextualized features are proven effective and critical to text representation for various NLP tasks \cite{ELMo,BERT}. For KBC, the features are significant for entity disambiguation. 
Therefore, despite slightly improving generalization, they still deliver an inferior performance. 
In contrast, our model achieves a better trade-off between deep contextualized features and efficiency by the carefully designed triple encoder. 

\paragraph{Joint Embedding.} 
More similar to our work, some other approaches \cite{xiao2017ssp,wang2014knowledge,xie2016representation,dieudonat2020exploring,xu2017TEKE,yamada2016joint,FB15k-237} also bring text representation learning into graph embedding paradigm. Standing opposite our model, they start with graph embeddings and aim at enriching the embeddings with text representations. 
Typically, they either use text embeddings to represent entities/relations and align heterogeneous representations into the same space \cite{wang2014knowledge,yamada2016joint,xie2016representation}, or employ large-scale raw corpora containing co-occurrence of entities to enrich the graph embeddings \cite{xu2017TEKE}. 
However, due to graph embeddings involved, they inevitably inherit the generalization problem and incompleteness issue. And same as the above, the representation learning here is also based on shallow networks without deep contextualized knowledge. 
In contrast, our model, based solely on text's contextualized representations and coupled with structure learning, is able to achieve mutual benefits of the two paradigms for KBC.

\section{Experiment} \label{sec:exp}

\begin{table}[t]\small
\caption{\small Summary statistics of benchmark datasets.}
\renewcommand\tabcolsep{4.0pt}
	\centering
	\begin{tabular}{c|ccccc}
		\hline
		\textbf{Dataset}   & \textbf{\# Ent}  & \textbf{\# Rel} & \textbf{\# Train} & \textbf{\# Dev} & \textbf{\# Test} \\ \hline
		WN18RR    & 40,943  & 11    & 86,835   & 3,034  & 3,134   \\
		FB15k-237 & 14,541  & 237   & 272,115  & 17,535 & 20,466  \\ 
		UMLS      & 135    & 46    & 5,216    & 652   & 661    \\ \hline
		NELL-One  & 68,545  & 822   & 189,635  & 1,004  & 2,158 \\\hline
	\end{tabular}
	\label{tb:benchmark_stat}
\end{table}

In this section\footnote{The source code is available at \url{https://github.com/wangbo9719/StAR_KGC}.}, we evaluate StAR on several popular benchmarks (\S \ref{subsec:main_evaluation}), and verify the model's efficiency (\S \ref{subsec:comp_baseline}) and generalization (\S \ref{subsec:exp_generalization}). 
Then, we conduct an extensive ablation study in \S \ref{subsec:ablation} to test various model selections and verify the significance of each proposed module. 
Lastly, in \S \ref{subsec:further_analyses} we comprehensively analyze the difference between graph embedding approach and textual encoding approach, and assess the self-adaptive ensemble scheme.

\begin{table*}[t]
\caption{\small Link prediction results on WN18RR, FB15k-237 and UMLS. 
		$\dagger$Resulting numbers are reported by \citet{nathani2019attention-based}, $\diamondsuit$Resulting numbers are re-evaluated by \cite{re-evaluation}, and others are taken from the original papers; UMLS results are reported by \citet{yao2019kgbert}, except ConvE from our re-implementation. 
		The bold numbers denote the best results in each genre while the underlined ones are state-of-the-art performance. 
	}
    \setlength{\tabcolsep}{3pt}
	\centering
	\begin{tabular}{l|ccccc|ccccc|cc}  \hline
	        & \multicolumn{5}{c|}{\textbf{WN18RR}}           & \multicolumn{5}{c|}{\textbf{FB15k-237}}    & \multicolumn{2}{c}{\textbf{ULMS}}      \\ \cline{2-6} \cline{7-11} \cline{12-13} 
	        & Hits@1   & Hits@3   & Hits@10           & MR          & MRR  & Hits@1     & Hits@3     & Hits@10      & MR  & MRR & Hits@10 & MR \\ \hline \hline\hline
		\multicolumn{7}{l}{\textit{Graph embedding approach}}\\ \hline
		TransE \cite{TransE}$\dagger$   & .043   & .441   & .532          & 2300        & .243 & .198    &.376   &.441   &323    &.279 & .989 & 1.84 \\  
		DistMult \cite{yang2014DistMult}$\dagger$ & .412   & .470   & .504          & 7000        & .444 & .199    &.301   &.446   &512    &.281 & .846 & 5.52 \\
		ComplEx \cite{ComplEx}$\dagger$ & .409   & .469   & .530          & 7882        & .449 & .194    &.297   &.450   &546    &.278 & .967 & 2.59 \\ 
		KBGAN \cite{cai2017kbgan}        & -      & -      & .469          & -           & .215 & -      & -      & .458    & -   & .277  & - & - \\
		R-GCN \cite{R-GCN}$\dagger$  & .080   & .137   & .207  & 6700    & .123 & .100    &.181   &.300   &600    &.164 & - & - \\
		ConvE \cite{ConvE}$\dagger$     & .419   & .470   & .531          & 4464        & .456 & .225    & .341  & .497  &245   & .312   & \textbf{.990} & \textbf{1.51} \\
		ConvKB \cite{nguyen2017ConvKB}$\diamondsuit$ & - & - & .524 &3433 &.249 & - & - &.421 &309 &.243 & - & - \\ 
		KBAT \cite{nathani2019attention-based}$\diamondsuit$    & -  & - &.554 &1921 &.412   & - & - &.331 &270 &.157 & - & -\\
		CapsE \cite{vu2019CapsE}$\diamondsuit$ & - & - &.559 &\textbf{718} &.415 & - & - &.356 &403 &.150 & - & -\\ 
		QuatE \cite{QuatE} &.436 &\textbf{.500} &.564 &3472 &.481    &.221 &.342 &.495 &\textbf{176} &.311 &- &- \\
		RotatE \cite{sun2019rotate}      & .428   & {.492}   & .571          & 3340        & .476 & .241   & .375   & .533    &{177} & .338 & - & - \\
		TuckER \cite{balavzevic2019tucker}&\textbf{.443} &.482 &.526 &- &.470   &\underline{\textbf{.266}} &\textbf{.394} &\textbf{.544} &- &\textbf{.358} & - & - \\
		\textsc{AttH}\cite{Chami2020ATTH} &\textbf{.443} &.499 &\textbf{.573} &- &\textbf{.486}    &.252 &.384 &.540 &- &.348 \\

		\hline  \hline \hline
		\multicolumn{7}{l}{\textit{Textual encoding approach}}\\ \hline
		KG-BERT \cite{yao2019kgbert}     & .041      & .302      & .524          & 97          & .216    & -      & -      & .420    & 153 & - & .990 &\underline{\textbf{1.47}} \\
		\textbf{StAR} &\textbf{.243} &\textbf{.491} &\textbf{.709} &\textbf{51} &\textbf{.401}      &\textbf{.205} &\textbf{.322} &\textbf{.482} &\textbf{117} &\textbf{.296}   & \textbf{.991} & 1.49 \\ 
		\hline \hline \hline
		\multicolumn{13}{l}{\textit{Our ensemble model}}\\ \hline
 		StAR (Ensemble)  & .449  &.551   &.675   &540 &.524       &.264    &.399  &.559   & \underline{\textbf{109}}    &.362       & - & - \\
		\textbf{StAR (Self-Adp)} &\underline{\textbf{.459}} &\underline{\textbf{.594}} &\underline{\textbf{.732}} &\underline{\textbf{46}} &\underline{\textbf{.551}}   &\underline{\textbf{.266}} &\underline{\textbf{.404}} &\underline{\textbf{.562}}   & 117 &\underline{\textbf{.365}}   &- &- \\ \hline
	\end{tabular}
	\label{tab:main_results}
\end{table*}
\paragraph{Benchmark Datasets.}
We assessed the proposed approach on three popular and one zero-shot link prediction benchmarks, whose statistics are listed in Table \ref{tb:benchmark_stat}. 
First, \emph{WN18RR} \cite{ConvE} is a link prediction dataset from WordNet \cite{miller1998wordnet}. It consists of English phrases and their semantic relations. 
Second, \emph{FB15k-237} \cite{FB15k-237} is a subset of Freebase \cite{db/freebase}, consisting of real-world named entities and their relations. 
Note, WN18RR and FB15k-237 are updated from WN18 and FB15k \cite{TransE} respectively by removing inverse relations and data leakage, which are the most popular benchmarks.\footnote{
WN18 and FB15k suffer from informative value \cite{FB15k-237,ConvE}, which causes $>80\%$ of the test triples $(e^1, r^1, e^2)$ can be found in the training set with another relation: $(e^1, r^2, e^2)$ or $(e^2, r^2, e^1)$. \citet{ConvE} used a rule-based model that learned the inverse relation and achieved state-of-the-art results on the dataset. Thereby it is suggested they should not be used for link prediction evaluation anymore. 
}
And third, \emph{UMLS} \cite{ConvE} is a small KG containing medical semantic entities and their relations. 
Finally, to verify model's generalization, \emph{NELL-One} \cite{one-shot} is a few-shot link prediction dataset derived from NELL \cite{NELL}, where the relations in dev/test set never appear in train set. We adopted ``In-Train'' scheme by \citet{few-shot} and used zero-shot setting. 
And, in line with prior approaches \cite{xiao2017ssp,yao2019kgbert}, we employed entity descriptions as their text for WN18RR and FB15k-237 from synonym definitions and Wikipedia paragraph \cite{xie2016representation} respectively. As for the text of relations and other datasets' entities, we directly used their text contents. 
Please refer to Appendix~\ref{app:train_setup} for our training setups.

\begin{table}[t] \small
\caption{\small Comparisons with KG-BERT on WN18RR. ``T/Ep'' stands for time per training epoch and ``Infer'' denotes inference time on test set. The time was collected on RTX6000 with mixed precision. }
    \setlength{\tabcolsep}{1pt}
	\centering
	\begin{tabular}{lccccc|cc}
		\hline
		& \textbf{Hits@1}      & \textbf{@3}      & \textbf{@10}    & \textbf{MR} & \textbf{MRR} & \textbf{T/Ep} & \textbf{Infer} \\ \hline
		KG-BERT$_{\text{BERT-base}}$& .041        &.302      & .524    &\underline{97}     & .216 & 40m & 32h \\ 
		\textbf{StAR$_{\text{BERT-base}}$}      &\underline{.222}	&\underline{.436}	&\underline{.647}	&99	&\underline{.364}  & 20m & 0.9h \\ \hline
		KG-BERT$_{\text{RoBERTa-base}}$  &.130        &.320       & \underline{.636}    &84     &.278  & 40m & 32h  \\ 
		\textbf{StAR$_{\text{RoBERTa-base}}$} &\underline{.202}	  &\underline{.410}	   & .621          & \underline{71}	   &\underline{.343}  & 20m & 0.9h  \\  \hline
		KG-BERT$_{\text{RoBERTa-large}}$  &.119        &.387       &.698  &95  & .297  & 79m & 92h \\
		\textbf{StAR$_{\text{RoBERTa-large}}$}   & \underline{.243}   & \underline{.491}   & \underline{.709} & \underline{51}    & \underline{.401}  & 55m & 1.0h \\ \hline
	\end{tabular}
	\label{tb:comp_baseline}
\end{table}

\paragraph{Evaluation Metrics.}
In the inference phase, given a test triple of a KG as the correct candidate, all other entities in the KG act as wrong candidates to corrupt its either head or tail entity. The trained model aims at ranking correct triple over corrupted ones with ``\textit{filtered}'' setting \citep{TransE}. 
For evaluation metrics, there are two aspects: (1) Mean rank (MR) and mean reciprocal rank (MRR) reflect the absolute ranking; and (2) Hits@$N$ stands for the ratio of test examples whose correct candidate is ranked in top-$N$. 
And, although there are two ranking scores from Eq.(\ref{eq:cls_score}) and Eq.(\ref{eq:dist_measure}), only $s^c$ is used for ranking, and other options will be discussed in \S\ref{subsec:ablation}. 

\paragraph{Evaluation Protocol.} We must emphasize that, as stated by \citet{re-evaluation}, previous methods (e.g., ConvKB, KBAT and CapsE) use an inappropriate evaluation protocol and thus mistakenly report very high results. 
The mistake frequently appears in a method whose score is normalized, says $[0,1]$, due to float precision problem. 
So, we strictly follow the ``RANDOM'' protocol proposed by \citet{re-evaluation} to evaluate our models, and avoid comparisons with vulnerable methods that have not been re-evaluated. 

\subsection{Evaluations on Link Prediction} \label{subsec:main_evaluation}

The link prediction results of competitive approaches and ours on the three benchmarks are shown in Table \ref{tab:main_results}. It is observed our proposed StAR is able to achieve state-of-the-art or competitive performance on all these datasets. The improvement is especially significant in terms of MR due to the great generalization performance of textual encoding approach, which will be further analyzed in the section below. 
And on WN18RR, StAR surpasses all other methods by a large margin in terms of Hits@$10$. 
Although it only achieves inferior performance on Hits@$1$ compared to graph embedding approaches, it still remarkably outperforms KG-BERT from the same genre by introducing structured knowledge. 

Further, coupled with the proposed self-adaptive scheme, the proposed model delivers new state-of-the-art performance on all metrics. 
Specifically, our self-adaptive model ``StAR (Self-Adp)'' significantly surpasses its ensemble baseline ``StAR (Ensemble)'' on most metrics. And, even if Hits@$1$ is the main weakness for a textual encoding paradigm, our self-adaptive model is still superior than the best semantic matching graph embedding approach TuckER.

\begin{table}[t] \small
\caption{\small Link prediction results on NELL-One. StAR with zero-shot setting is competitive with few-shot GMatching \cite{one-shot} and MetaR \cite{few-shot}.
	} 
    \setlength{\tabcolsep}{1pt}
	\centering
	\begin{tabular}{l|c|cccc}
		\hline
		\textbf{Methods}&  \textbf{$\bm N$-Shot}  &\textbf{Hits@1}   & \textbf{Hits@5}    & \textbf{Hits@10}   &\textbf{MRR} \\ \hline
		GMatching$_{\text{ComplEx}}$  &\multirow{2}{*}{Five-Shot} &.14 &.26 &.31 &.20 \\
		MetaR  & ~ &.17 &.35 &.44 &.26 \\ \hline
 		GMatching$_{\text{TransE}}$ &\multirow{4}{*}{One-Shot} &.12 &.21 &.26 &.17 \\
		GMatching$_{\text{DistMult}}$ & ~ &.11 &.22 &.30 &.17 \\
		GMatching$_{\text{ComplEx}}$ & ~ &.12 &.26 &.31 &.19 \\
		MetaR  & ~ &.17 &.34 &.40 &.25 \\ \hline
		\textbf{StAR$_{\text{BERT-base}}$} & Zero-Shot&\textbf{.17} &\textbf{.35} &\textbf{.45} &\textbf{.26} \\ \hline
		
	\end{tabular}
	\label{tb:generalization_rel}
\end{table}

\subsection{Comparison with KG-BERT Baseline} \label{subsec:comp_baseline}

Since our approach is an update from the non-Siamese-style baseline, says KG-BERT, we compared StAR with KG-BERT on WN18RR in detail, including different initializations. As shown in Table \ref{tb:comp_baseline}, our proposed StAR consistently achieves superior performance over most metrics. 
As for empirical efficiency, it is observed our model is faster than KG-BERT despite training or inference, which is roughly consistent with the theoretical analysis in \S\ref{subsubsec:efficiency}.

\subsection{Generalization to Unseen Graph Elements} \label{subsec:exp_generalization}

Textual encoding approaches are more generalizable to unseen entities/relations than graph embedding ones. 
This can be more significant when the set of entities or relations is not closed, i.e., unseen graph elements (i.e., entities/relations) appear during inference. For example, 209 out of 3134 and 29 out of 20466 test triples involve unseen entities on WN18RR and FB15k-237 respectively. This inevitably hurts the performance of graph embedding approaches, especially for the unnormalized metric MR.

First, we employed a few-shot dataset, NELL-One, to perform a zero-shot evaluation where relations in test never appear in training set. As shown in Table \ref{tb:generalization_rel}, StAR with zero-shot setting is competitive with graph embedding approaches with one/five-shot setting. 

\begin{table}[t] \small
	\caption{\small Probing tasks based on WN18RR for analyzing models' generalization performance. } 
    \setlength{\tabcolsep}{1pt}
	\centering
	\begin{tabular}{clccccc}
		\hline
		\multicolumn{1}{l}{}    &        & \textbf{Hits@1} & \textbf{Hits@3} & \textbf{Hits@10} & \textbf{MR}    & \textbf{MRR}  \\ \hline
		\multicolumn{1}{c|}{\multirow{3}{*}{\begin{tabular}[c]{@{}c@{}}Original Task\end{tabular}}}  & \textbf{StAR}    & .243   & .491   & .709    & 51    & .401 \\
		\multicolumn{1}{c|}{}  & RotatE & .428   & .492   & .571    & 3340  & .476 \\
		\multicolumn{1}{c|}{}  & TransE & .042   & .441   & .532    & 2300  & .243 \\ \hline
		\multicolumn{1}{c|}{\multirow{3}{*}{\begin{tabular}[c]{@{}c@{}}First\\ Probing Task\end{tabular}}} & \textbf{StAR}  & .240   & .452   & .673    & 71    & .384 \\
		\multicolumn{1}{c|}{}  & RotatE & .005   & .007   & .012    & 17955 & .007 \\
		\multicolumn{1}{c|}{}  & TransE & .000      & .007       &.016         & 20721      & .007     \\ \hline
		\multicolumn{1}{c|}{\multirow{2}{*}{\begin{tabular}[c]{@{}c@{}}Second\\ Probing Task \end{tabular}}}  & \textbf{StAR}    & .301   & .497   & .676    & 99    & .427 \\
		\multicolumn{1}{c|}{}  & TransE & .005   & .121   & .210     & 13102 & .078 \\ \hline
		\multicolumn{1}{c|}{\multirow{2}{*}{\begin{tabular}[c]{@{}c@{}}Third\\ Probing Task \end{tabular}}}  & \textbf{StAR}    & .244   & .493   & .712    & 49    & .402 \\
		\multicolumn{1}{c|}{}  & RotatE & .455   & .523   & .612     & 1657 & .507 \\ \hline
	\end{tabular}
	\label{tb:generalization}
\end{table}

Then, to verify the generalization to unseen entities, we built two probing settings on WN18RR. The \emph{first probing task} keeps training set unchanged and makes the test set only consist of the triples with unseen entities. And in the \emph{second probing task}, we randomly removed 1900 entities from training set to support inductive entity representations \citep{hamilton2017inductive} during test for TransE. The setting is detailed in Appendix~\ref{app:two_probing}. 
As shown in Table \ref{tb:generalization}, StAR is competitive across the settings but advanced graph embedding approaches (e.g., RotatE) show a substantial drop in the first task. Even if we used translation formula to inductively complete unseen entities' embeddings in the second probing task, the degeneration of TransE is significant. 
These verify StAR's promising generalization to unseen elements.

Lastly, to verify the proposed model is still competitive even if applied to close sets of entities/relations, we built the \textit{third probing task} as in Table~\ref{tb:generalization}. We only kept the WN18RR test triples with entities/relations visited during training while removed the others.

\subsection{Ablation Study}  \label{subsec:ablation}

\begin{table}[t] \small
    \setlength{\tabcolsep}{1pt}
    \caption{\small Ablation study on WN18RR. 
    Note that $^*$Full model denotes using two objectives for training, `` [\textit{h, r}] vs. [\textit{t}]''  as concatenation scheme, $L2$ norm as measurement, and $s^c$ as ranking basis during inference. 
    And $\rescale(\cdot)$ denotes scaling all scores to $[0, 1]$. 
    }
	\centering
	\setlength{\tabcolsep}{2.5pt}
	\begin{tabular}{llccc}
		\hline
		\multicolumn{1}{l}{\textbf{Perspective}}               &  \textbf{Detail}              & \textbf{Hits@10}    & \textbf{MR} & \textbf{MRR}  \\ \hline
		\hline
		\multicolumn{5}{l}{\textit{\textbf{Single Model}: Module Ablation and Selection in ``StAR''}} \\
		\hline
		\textit{Full model}$^*$    &StAR$_{\text{RoBERTa-large}}$       & {.709} & {51}    & {.401} \\ 
		\hline
		\multirow{2}{*}{\textit{Objective}} &$\cdot$ w/o contrastive obj     & .685   & 68 & .399 \\
		&$\cdot$ w/o classification obj      & .653   & 67 & .337 \\ 
		\hline
		\multirow{1}{*}{\textit{Concatenation,}} 
		& $\cdot$ [\textit{h, r}] vs. [\textit{r, t}]  &.520 &106 &.204 \\  
		\multirow{1}{*}{e.g., Eq.(\ref{eq:cat_left}, \ref{eq:cat_right})} & $\cdot$ [\textit{h}] vs. [\textit{r, t}]  &.668 &51 &.402 \\ 
		\hline
		\multirow{1}{*}{\textit{Distance}} 
		&$\cdot$ Bilinear     &.605 &79 &.354 \\ 
		\multirow{1}{*}{\textit{in} Eq.(\ref{eq:dist_measure})} &$\cdot$ Cosine Similarity       &.691 &76 &.439 \\ 
		\hline
		\multirow{3}{*}{\textit{Ranking Basis}} 
		&$\cdot$ $s^d$    	&.701 &62 &.406 \\
		&$\cdot$ $\rescale(s^d) + s^c$  	&.706	&48	&.408 \\
		&$\cdot$ $\rescale(s^d) \times s^c$	&.704	&51	 &.408 \\ \hline
		\hline
		\multicolumn{5}{l}{\textit{\textbf{Ensemble Model}: Feature Ablation in ``StAR (Self-Adp)''}} \\
		\hline
		\multirow{1}{*}{\textit{Full model}} 
		&   StAR (Self-Adp)	&.732 &46 &.551 \\
		\hline
		\multirow{4}{*}{\textit{Feature}} 
		&$\cdot$ w/o hard indicator   	&.712 &50 &.540 \\
		&$\cdot$ w/o ambiguity degree  	&.734	&45	&.537 \\
		&$\cdot$ w/o score consistency &.720	&45	 &.540 \\ 
		&$\cdot$ w/o self-Adp $\alpha$ in Eq.(\ref{eq:self_adp_alpha}) &.675	&540	 &.524 \\ \hline
	\end{tabular}
	\label{tb:ablation}
\end{table}

To explore each module's contribution, we conducted an extensive ablation study about StAR and the self-adaptive ensemble scheme as shown in Table \ref{tb:ablation}. 
For single StAR, 
(1) \emph{Ablating Objective}: First, each of the components in Eq.(\ref{eq:loss_total}) were removed to estimate the significance of structure and representation learning. 
(2) \emph{Contexts' concatenation}: 
Then, how to concatenate and encode the text from a triple is also non-trivial for learning structured knowledge. Two other options only achieved sub-optimal results. 
(3) \emph{Distance measurement}: Two other methods, i.e., Bilinear and Cosine, similarity were also applied to Eq.(\ref{eq:dist_measure}) to measure the distance for structure learning. 
(4) \emph{Ranking Basis}: Since two scores can be derived from the two objectives respectively, it is straightforward to integrate them in either additive or multiplicative way. As these ranking bases achieve similar performance, we further calculated the Pearson correlation between $s^c$ and $s^d$ and found the coefficient is $0.939$ (p-value=$7\!\times\!10^{-4}$), which means the two scores are linearly related. For ``StAR (Self-Adp)'' (in \S~\ref{subsec:adaptive model}), we ablated its features: (1) unseen indicator, (2) ambiguity degree, and (3) score consistency.

\subsection{Further Analyses} \label{subsec:further_analyses}
Here, we analyze the effect of structure learning on our textual encoding approach, and compare the proposed model with a graph embedding approach. 
And we also attempt to qualitatively measure the effectiveness of the proposed self-adaptive ensemble scheme.

\begin{figure}[t] \centering
    \includegraphics[width=0.48\textwidth]{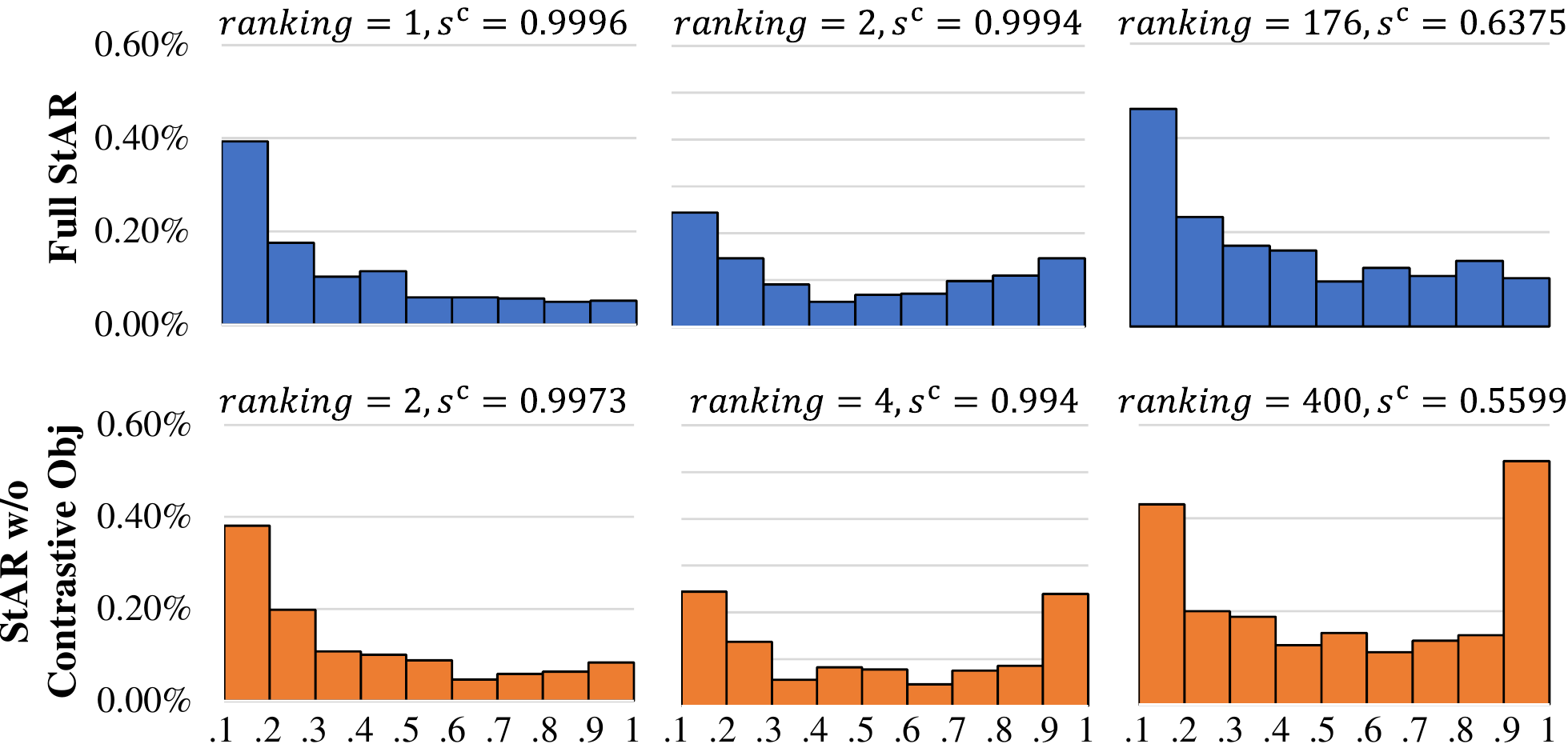}
	\caption{\small Three random comparative cases of frequency histogram for $s^c~\!'$ assigned to a triple's all tail corruptions. $x$-axis denotes $s^c~\!'$ and $y$-axis denotes frequency over the number of all corruptions. 
	The text above each histogram shows the ranking and $s^c$ for the corresponding un-corrupted (i.e., oracle) triple. 
	Note, interval of [0.0, 0.1] is removed since most negative triples' $s^c~\!'$ will fall into it. 
	} 
	\label{fig:freq} 
\end{figure} 

\begin{table*}[t]
\center
\setlength{\tabcolsep}{3pt}
\caption{\small Top-5 ranking results of candidate entities for different approaches. The first column includes incomplete triples for inference, and their labels. And the others include the ranking position and Top-5 ranked candidates where an underline denotes it is the gold entity.} 
\begin{tabular}{c|ccc}
\hline
\multicolumn{1}{c|}{\multirow{2}{*}{Incomplete Triple}} & \multicolumn{3}{c}{Positive entity ranking position \& Top-5 ranked candidate entities} \\ \cline{2-4} 
\multicolumn{1}{c|}{}                        & \multicolumn{1}{c|}{\textbf{StAR (Self-Adp)} [Ensemble]} & \multicolumn{1}{c|}{\textbf{StAR} [Textual Encoding]} & \textbf{RotatE} [Graph Embedding] \\ \hline
\multicolumn{1}{c|}{\multirow{2}{*}{\begin{tabular}[c]{@{}c@{}}(\textit{world war ii, has part,} ?) \\ $\leftarrow$ \textit{tarawa-makin} \end{tabular}}}
& \multicolumn{1}{c|}{\multirow{2}{*}{\begin{tabular}[c]{@{}c@{}}5, (\textit{world war ii, jutland, meuse river,}\\ \textit{soissons, \underline{tarawa-makin}}) \end{tabular}}}
& \multicolumn{1}{c|}{\multirow{2}{*}{\begin{tabular}[c]{@{}c@{}}12, (\textit{world war ii, world war i, world}\\ \textit{war, seven years' war, meuse river}) \end{tabular}}}  
& \multirow{2}{*}{\begin{tabular}[c]{@{}c@{}}10, (\textit{jutland, world war ii,}\\ \textit{somme river, verdun, soissons}) \end{tabular}} \\
                                             & \multicolumn{1}{l|}{}                 & \multicolumn{1}{l|}{}      &        \\ \hline
\multicolumn{1}{c|}{\multirow{2}{*}{\begin{tabular}[c]{@{}c@{}}(\textit{clarify, hypernym,} ?) \\ $\leftarrow$ \textit{modify} \end{tabular}}}
& \multicolumn{1}{c|}{\multirow{2}{*}{\begin{tabular}[c]{@{}c@{}}2, (\textit{clarify, \underline{modify}, change integrity,}\\ \textit{$\textit{convert}^a$, $\textit{convert}^b$}) \end{tabular}}}
& \multicolumn{1}{c|}{\multirow{2}{*}{\begin{tabular}[c]{@{}c@{}}3, (\textit{clarify, straighten out, \underline{modify}, }\\ \textit{alter, transubstantiate}) \end{tabular}}}  
& \multirow{2}{*}{\begin{tabular}[c]{@{}c@{}}66, (\textit{$\textit{cook}^a$, season, ready,}\\ \textit{$\textit{cook}^b$, preserve}) \end{tabular}} \\
                                             & \multicolumn{1}{l|}{}                 & \multicolumn{1}{l|}{}      &        \\  [2pt] \hline
\multicolumn{1}{c|}{\multirow{2}{*}{\begin{tabular}[c]{@{}c@{}}(\textit{mechanical system,}\\ \textit{ hypernym, ?})  $\leftarrow$ \textit{$system^a$} \end{tabular}}}
& \multicolumn{1}{c|}{\multirow{2}{*}{\begin{tabular}[c]{@{}c@{}}2, (\textit{mechanical system, \underline{$system^a$},}\\ \textit{mechanism, $system^b$, machine)} \end{tabular}}} 
& \multicolumn{1}{c|}{\multirow{2}{*}{\begin{tabular}[c]{@{}c@{}}3, (\textit{$system^b$, mechanical system,}\\ \textit{\underline{$system^a$},mechanism, $system^c$)}  \end{tabular}}}  
& \multirow{2}{*}{\begin{tabular}[c]{@{}c@{}}24, (\textit{mechanical system, production}\\ \textit{line, suspension system, $\dots$})
\end{tabular}} \\
                                             & \multicolumn{1}{l|}{}                 & \multicolumn{1}{l|}{}      &        \\ [3pt] \hline
\end{tabular}
\label{tb:improvement}
\end{table*}

\begin{figure}[t]
	\centering
	\includegraphics[width=0.47\textwidth]{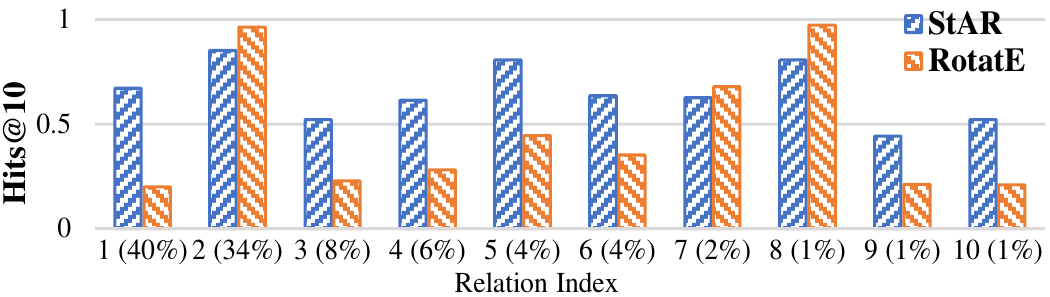}
	\caption{\small A comparison between StAR and RotatE regarding different relations on WN18RR test set. 
	Relations corresponding to the indices are 
	1) \textit{hypernym}, 2) \textit{derivationally related form}, 3) \textit{member meronym}, 
	4) \textit{has part}, 5) \textit{instance hypernym}, 6) \textit{synset domain topic of}, 
	7) \textit{also see}, 8) \textit{verb group}, 9) \textit{member of domain region}, 
	10) \textit{member of domain usage}. 
	The number in a parenthesis denotes its proportion of test triples with the corresponding relation. Note relation ``\textit{similar to}'' is ignored since its proportion is less than 0.1\%.
	}
	\label{fig:rel_results} 
	\centering
\end{figure}

\paragraph{What is the effect of introducing structure learning into a textual encoding approach.}

As shown in Figure \ref{fig:freq}, 
we compared the frequency histograms of ranking score $s^c~\!'$ derived from classification objective for negative triples from either the ``full StAR'' or ``StAR w/o contrastive objective''. 
It is observed, the textual encoding model augmented with structure learning reduces the number of false positive predictions and produces more accurate ranking scores. This also verifies the structured knowledge can alleviate the over-confidence problem (\S\ref{subsubsec:train_infer_details}).

\paragraph{How does StAR bring improvements.} 
As shown in Figure \ref{fig:rel_results}, a detailed comparison regarding different relations is conducted between StAR and RotatE. It is observed StAR achieves more consistent results than RotatE. 
However, StAR performs worse on several certain relations, e.g., the $8th$ relation in Figure \ref{fig:rel_results}, \textit{verb group}. After checking the test triples falling into \textit{verb group}, we found ``polysemy'' occurs in half of the triples, e.g., (\textit{strike$^a$, verb group, strike$^b$}), which hinders StAR from correctly ranking. 
These imply that even coupled with structured knowledge, a textual encoding approach is still vulnerable to entity ambiguity or word polysemy, and emphasize the importance of our self-adaptive ensemble scheme. 

\paragraph{Why does StAR achieve better Hits@10 but worse Hits@1 than RotatE}
As shown in Table \ref{tab:main_results}, it is observed that the textual encoding approach (e.g., KG-BERT, StAR) can outperform graph embedding approach (e.g., TransE, RotatE) by a large margin on Hits@10 but underperform on Hits@1. 
To dig this out, we conducted a case study based on the inference on WN18RR. 
In particular, given an oracle test triple, (\textit{sensitive$^{a}$, derivationally related form, \underline{sense}}), after corrupting its tail and ranked by our StAR, the top-12 tail candidates are (\textit{sensitive$^{a}$, sensitivity, sensibility, sensing, sense impression, sentiency, sensitive$^{b}$, \underline{sense}, feel, sensory, sensitive$^{c}$, perceptive}), where gold tail is only ranked $8th$. It is observed there are many semantically-similar tail entities that can fit the oracle triple, which seem to be false negative labels for a context encoder.
But this is not a matter for graph embedding approaches since they only consider graph's structure despite text. 
It is worth mentioning ``polysemy'' or ``ambiguity'' issue usually appears in WN18RR (an example in Table \ref{tb:example}). The issue is more severe in FB15K-237, which partially explains why StAR only achieves competitive results. 
Fortunately, this issue can be significantly alleviated by the self-adaptive ensemble scheme. 
And, it is interesting the oracle head is ranked $1st$ for tail in this case but self-loop will never appear in WN18RR's test set. Hence, as shown in Table \ref{tb:self_loop}, after filtering self-loop candidates during inference, the performance is improved.

\begin{table}[t] \small
    \caption{\small An example of polysemy in WordNet: three meanings of ``\textit{sensitive}'' are viewed as three separate nodes.}
	\centering
	\begin{tabular}{p{0.9\columnwidth}}
		\hline
		$\bullet$ sensitive$^{a}$: able to feel or perceive. \\
		$\bullet$ sensitive$^{b}$: responsive to physical stimuli. \\
		$\bullet$ sensitive$^{c}$: being susceptible to the attitudes, feelings, or circumstances of others. \\
		\hline
	\end{tabular}
	\label{tb:example}
\end{table}

\begin{table}[t] \small
    \caption{\small Applying self-loop filter to WN18RR. }
    \setlength{\tabcolsep}{4pt}
	\centering
	\begin{tabular}{lccccc}
		\hline
		& \textbf{Hits@1}      & \textbf{@3}      & \textbf{@10}    & \textbf{MR} & \textbf{MRR}  \\ \hline
		StAR & .243   & .491   & .709    & 51    & .401  \\ 
		~~+ Self-loop Filter  &.328	&.533	&.719	&50	&.460   \\ \hline
	\end{tabular}
	\label{tb:self_loop}
\end{table}

\paragraph{How does the self-adaptive ensemble scheme bring improvements.}
As shown in Table \ref{tab:main_results}, ``StAR (Self-Adp)'' improves the performance than ``StAR'' or RotatE used alone. 
Intuitively, the improvement is brought from the mutual benefits of representation and structure learning. 
For further confirmation, we randomly listed some triples in WN18RR test, where the triples experience a certain improvement when applying self-adaptive ensemble scheme. 
As shown in Table \ref{tb:improvement}, as demonstrated in the $1st$ and $3rd$ examples, it is observed that graph structure helps distinguish semantically-similar candidate entities and alleviate the "polysemy" problem. 
In addition, since the rich contextualized information empowers model with a high top-$k$ recall, the self-adaptive ensemble model still achieves a satisfactory ranking result as shown in the $2nd$ example, even if the graph embedding model underperforms. 
As a result, due to the complementary benefits, the self-adaptive ensemble scheme offers significant improvements over previous approaches.

\section{Related Work} \label{sec:related}
\paragraph{Structure Learning for Link Prediction.}
Previous graph embedding approaches explore structured knowledge through spatial measurement or latent matching in low-dimension vector space. 
Specifically, on the one hand, translation-based graph embedding approach \cite{TransE,sun2019rotate} applies a translation function to \textit{head} and \textit{relation}, and compares the resulting with \textit{tail} via spatial measurement. 
The most well-known one, TransE \cite{TransE}, implements the function and the measurement with real vector addition and $L2$ norm respectively -- scoring a triple by $-||(\vh + \vr) - \vt||$. 
However, the graph embeddings defined in real vector space hardly deal with the \textit{symmetry} relation pattern, and thereby underperform. 
To remedy this, RotatE \cite{sun2019rotate} defines the graph embeddings in complex vector space, and implements the translation function with the production of two complex numbers in each dimension. 
On the other hand, semantic matching graph embedding approach \cite{yang2014DistMult,balavzevic2019tucker,QuatE} uses a matching function $f(\vh,\vr,\vt)$ operating on whole triple to directly derive its plausibility score. 
For example, DistMult \cite{yang2014DistMult} applies a bilinear function to each triple's components and uses the latent similarity in vector space as the plausibility score. 
In spite of their success, the rich text contextualized knowledge is entirely ignored, leading to less generalization. 

\paragraph{Text Representation Learning.}
In an NLP literature, text representation learning is fundamental to any NLP task, which aims to produce expressively powerful text embedding with contextualized knowledge \cite{ELMo,BERT}. 
When applied to KGC, some approaches \cite{socher2013kgc,McIlraith2018open} directly replace the graph embeddings with their text embedding. 
For example, \citet{socher2013kgc} simply use continuous CBoW as the representation of triple's component, and then proposed a neural tensor network for relation classification. 
ConMask \cite{McIlraith2018open} learns relationship-dependent entity embeddings of the entity’s name and parts of description based on fully CNN. 
These approaches are not competitive since the deep contextualized representation of a triple is not leveraged. 
In contrast, KG-BERT \cite{yao2019kgbert}, as a textual encoding approach, applies pre-trained encoder to a concatenation of triples' text for deep contextualized representations. Such a simple method is very effective, but unfortunately suffers from high overheads.

\paragraph{Jointly Learning Methods.}
Unlike the approaches above learning either knowledge solely, several works explore jointly learning both text and structured knowledge. 
Please refer to the end of \S \ref{subsec:compared_text-based} for more detail. 
For example, taking into account the sharing of sub-structure in the textual relations in a large-scale corpus, \citet{FB15k-237} applied a CNN to the lexicalized dependency paths of the textual relation, for augmented relation representations. 
\citet{xie2016representation} propose a representation learning method for KGs via embedding entity descriptions, and explored CNN encoder in addition to CBoW. They used the objective across this representation and graph embeddings that a vector integration of head and relation was close to the vector of tail to learn the model, as in translation-based graph embedding approaches \cite{TransE}. 
In contrast, our work only operates on homogeneous textual data and employs the contexts for entities/relations themselves (i.e., only their own text contents or description), rather than acquiring textual knowledge (e.g., textual relations by \citet{FB15k-237}) from large-scale corpora to enrich traditional graph embeddings via joint embedding.

\section{Conclusion} \label{sec:conclusion}
In this work, we propose a structure-augmented text representation (StAR) model to tackle link prediction task for knowledge graph completion. 
Inspired by translation-based graph embedding designed for structure learning, we first apply a Siamese-style textual encoder to a triple for two contextualized representations. 
Then, based on the two representations, we present a scoring module where two parallel scoring strategies are used to learn both contextualized and structured knowledge. 
Moreover, we propose a self-adaptive ensemble scheme with graph embedding approach, to further boost the performance. 
The empirical evaluations and thorough analyses on several mainstream benchmarks show our approach achieves state-of-the-art performance with high efficiency. 

\section*{Acknowledgement}
The authors would like to thank the anonymous referees for their valuable comments. This work is supported by the National Natural Science Foundation of China (No.61976102, No.U19A2065, No.61872161) and the Foundation of Development and Reform of Jilin Province (2019C053-8).

\bibliographystyle{ACM-Reference-Format}
\bibliography{ref}

\appendix

\begin{table}[t] \small
\setlength{\tabcolsep}{2pt}
    \centering
    \caption{\small The grid searching of hyperparameters. Note, the hyperparameters in the first part, i.e., Batch Size and $\gamma$, were tuned based on WN18RR benchmark, RoBERTa initialization, learning rate $=10^{-5}$, number of training epochs $=6$. After the first was part tuned, the remaining was tuned subsequently. }
    \label{tab:hyperpar}
    \begin{tabular}{c|ccc} \hline
        \textbf{Hyperparm} &\textbf{Note}  &\textbf{Value}  &\textbf{Search scope} \\ \hline
        Batch Size     & -  &16     &\{16, 32, 64\} \\
        $|\gN(tp)|$    & -  &5      & \{5\} \\
        $\lambda$  & -      &1.0    & \{1.0\} \\ 
        $\gamma$   & -      &1.0    & \{0.5, 1.0, 2.0\} \\ \hline
        \multirow{2}{*}{Learning Rate}   & RoBERTa  &$10^{-5}$    &\multirow{1}{*}{\{$10^{-5}$\} } \\
        ~   & BERT   &$5\times10^{-5}$    &\multirow{1}{*}{\{$5\times10^{-5}$\}} \\ \hline
        ~    &WN18RR  & 7     & \multirow{3}{*}{\{6, 7, 8, 9\}} \\ 
        \multirow{1}{*}{Number of}   &FB15k-237 & 7 &~ \\
        \multirow{1}{*}{Training Epochs}   &NELL-One  &8  &~ \\ \cline{2-4}
        ~   &UMLS  &20 &\multirow{1}{*}{\{5-25\}}  \\ \hline
        
    \end{tabular}
\end{table}

\section{Training Setups} \label{app:train_setup}
In training phase, the initialization of Transformer encoder is alternated between BERT and RoBERTa. The model is fine-tuned by Adam optimizer. 
For the hyperparameters in StAR, based on the best Hits@10 on dev set, we set batch size $=16$, learning rate $=10^{-5}/5 \times 10^{-5}$ for the models initialized with RoBERTa and BERT respectively, number of training epochs $=7$ on WN18RR and FB15k-237, 8 on NELL-One, 20 on UMLS, $|\gN(tp)|=5$, $\lambda=1$ in Eq.(\ref{equ:hingeloss}), and $\gamma=1$ in Eq.(\ref{eq:loss_total}). As for grid searching of hyperparameters, we list the searching scopes and the tuned hyperparameters for best in Table \ref{tab:hyperpar}. 
Note, we sampled 5 negative triples for each positive triple by following \citet{yao2019kgbert} without any tuning, and we also did not tune the random seed while kept the same among the experiments.

For the hyperparameters in self-adaptive ensemble scheme, based on the best Hits@10 on WN18RR/FB15k-237 dev set, we set batch size $=32/64$, learning rate $=10^{-3}/10^{-5}$, number of training epochs $=1$, number of negative samples $=5/10$, and margin $=0.60/0.44$ in hinge loss function.

\section{Probing Tasks} \label{app:two_probing}

The \emph{first probing task} keeps training set unchanged but makes the test set only consist of the test triples involving unseen entities. And, in \emph{second probing task}, we conducted a more reasonable comparison by supporting inductive representations \citep{hamilton2017inductive} for unseen entities in a translation-based approach, and thus made following changes : (1) 1900 entities were sampled from test set, and only a test triple containing at least one of the sampled entities can be kept, resulting in 1758 test triples in this probing task; (2) Those training triples that do not contain the sampled entities are used as new training set; and (3) Those training triples containing exact one of the sampled entities are used as support set to inductively generate the embedding for the unseen entities via translation formula, such as ``$\vh+\vr = \vt$'' in TransE \citep{TransE}. 
Using the second probing setting can assign the unseen entities with competent embeddings, thus leading to a fairer comparison than the first one. 
Note, if an unseen entity is involved in multiple triple on the support set, an average over the multiple inductive representations is used as its single vector representation.\end{document}